\documentclass{article}

\usepackage{PRIMEarxiv}

\usepackage[utf8]{inputenc} 
\usepackage[T1]{fontenc}    
\usepackage{hyperref}       
\usepackage{url}            
\usepackage{booktabs}       
\usepackage{amsfonts}       
\usepackage{nicefrac}       
\usepackage{microtype}      
\usepackage{lipsum}
\usepackage{fancyhdr}       
\usepackage{graphicx}       
\usepackage{threeparttable}
\graphicspath{{media/}}     

\title{Millimeter Wave Radar-based Human Activity Recognition for Healthcare Monitoring Robot
}

\author{
  Zhanzhong Gu\\
  University of Technology Sydney \\
  \texttt{Zhanzhong.gu@student.uts.edu.au} \\
  \And
  Xiangjian He \\
  University of Nottingham Ningbo China \\
  \texttt{Sean.He@Nottingham.edu.cn} \\
  \AND
  Gengfa Fang \\
  The Hong Kong University of Science and Technology (Guangzhou) \\
  \texttt{gengfafang@hkust-gz.edu.cn} \\
  \And
  Chengpei Xu \\
  University of New South Wales \\
  \texttt{chengpei.xu@unsw.edu.au} \\
  \And
  Feng Xia \\
  RMIT University \\
  \texttt{feng.xia@rmit.edu.au} \\
  \And
  Wenjing Jia \\
  University of Technology Sydney \\
  \texttt{wenjing.jia@uts.edu.au} \\
}

\begin{document}
\maketitle

\begin{abstract}
Healthcare monitoring is crucial, especially for the daily care of elderly individuals living alone. It can detect dangerous occurrences, such as falls, and provide timely alerts to save lives. Non-invasive millimeter wave (mmWave) radar-based healthcare monitoring systems using advanced human activity recognition (HAR) models have recently gained significant attention. However, they encounter challenges in handling sparse point clouds, achieving real-time continuous classification, and coping with limited monitoring ranges when statically mounted.
To overcome these limitations, we propose RobHAR, a movable robot-mounted mmWave radar system with lightweight deep neural networks for real-time monitoring of human activities. Specifically, we first propose a sparse point cloud-based global embedding to learn the features of point clouds using the light-PointNet (LPN) backbone. Then, we learn the temporal pattern with a bidirectional lightweight LSTM model (BiLiLSTM). In addition, we implement a transition optimization strategy, integrating the Hidden Markov Model (HMM) with Connectionist Temporal Classification (CTC) to improve the accuracy and robustness of the continuous HAR. Our experiments on three datasets indicate that our method significantly outperforms the previous studies in both discrete and continuous HAR tasks. Finally, we deploy our system on a movable robot-mounted edge computing platform, achieving flexible healthcare monitoring in real-world scenarios.
\end{abstract}


\section{Introduction}
{W}{ith} the rapid development of artificial intelligence techniques~\cite{wang2019indoor,ali2023new,gu2022strokepeo,zhang2021rethinking,gu2024automatic}, the task of how to use the machine to automatically recognize human activities (HAR) has become a research hotspot in many application fields. It can be implemented in indoor monitoring scenarios to perform home behavior analysis~\cite{arab2022convolutional}, fall detection~\cite{jin2022mmfall} and intruder detection~\cite{ali2023new, rambabu2019iot}. 
In the field of healthcare monitoring, researchers apply HAR technologies for sleep and respiration detection~\cite{khan2017deep,zhao2020heart}, gait and abnormal behavior detection~\cite{hannink2016sensor}. 
Other areas such as sports analysis~\cite{hsu2018human}, and human-machine interaction~\cite{reily2018skeleton} also benefit from HAR technologies. 
The HAR research is primarily based on three categories of sensing devices: visual cameras, wearable devices, and contactless sensors. 
Although numerous efforts have been invested in visual-based HAR studies~\cite{singh2019human} and have achieved high recognition accuracy, optical cameras have some inherent flaws, such as being sensitive to lighting and blocking, and poor privacy protection. This has caused significant limitations in many specific scenarios such as restrooms and bathrooms~\cite{gurbuz2019radar}. 
Wearable devices such as smart watches and smart glasses mainly employ built-in accelerometers, magnetometers and gyroscopes to acquire motion information of the human body. 
However, due to the need to frequently charge and wear, people often forget to charge causing the device to shut down, or even forget to wear the device~\cite{lara2012survey}. 
Recently, healthcare monitoring robots with non-contact radar sensors have received extensive attention. 
Their immunity to illumination and blocking, as well as privacy-preserving features, make them superior to optical cameras and wearables in HAR applications. 
Moreover, in contrast to video data, radar data requires lower bandwidth due to small data volume~\cite{li2019survey,zhang2023survey}. 
This study indicates that contactless radar can realize real-time continuous HAR on robot-mounted edge devices and achieve superb classification results. 

This paper specifically focuses on a millimeter-wave (mmWave) radar. The mmWave radar works in the wavelength range between 1 and 10 mm. 
Thus, it has the following advantages compared to traditional sensing technologies. First, it can provide high-precision multi-dimensional search and measurement to perform high-precision distance, azimuth, frequency and spatial position measurement and positioning for moving objects$\footnote{http://www.ti.com/sensors/mmwave/overview.html.}$.  
Secondly, mmWave radar is more unobtrusive since it can be concealed inside furniture or walls as it can penetrate thin layers of materials~\cite{zhao2019mid}. 
Third, mmWave radar units can be low cost, low power, small in size, and can be mounted on walls and ceilings in any room~\cite{jin2022mmfall}, or be mounted on movable edge computing platforms as healthcare monitoring robot~\cite{zhao2020heart}.  

The data generated by mmWave radar mainly contains a 3D position (measured by the Range and the Azimuth) and a 1D Doppler (radial velocity component) information~\cite{jin2022mmfall}. 
Traditionally, most radar-based HARs are based on Range and Doppler data, which generate 1D to 3D feature maps from the Time-Range-Doppler domain information for HAR tasks~\cite{zhang2023survey}. 
However, the attributes of Range and Doppler data also cause them to perform poorly in some tasks. First, activities of similar motions, like falling compared to fast sitting and kneeling, as well as pets jumping off the table, have similar time-Doppler signatures which may cause false alarms~\cite{amin2016radar}. 
Secondly, radar is insensitive in the direction of orthogonality direction to the radar, where only a weak Doppler signature is received~\cite{amin2016radar}. 
Thirdly, radar-based human activity data are still insufficient, especially for occasional actions such as falls and crawls. This is because it is difficult to collect data of these rare events~\cite{jin2022mmfall}. 
Additionally, little evidence is found from previous literature about data augmentation applied to extend the original radar data. 
Thus, an increasing number of emerging studies tend to choose 3D point cloud data in mmWave radar-based HAR~\cite{zhao2019mid,jin2022mmfall,singh2019radhar,zhang2018real}. 
However, it is challenging to represent and learn the pattern based on sparse and imbalanced point clouds. 
An extensive number of existing studies implement 2D histogram features or 3D volumetric representations. 
One limitation is the trade-off between the granularity of feature extraction and exponentially increased computational intensity. 
In addition, most previous research is still limited to non-continuous data in a controlled laboratory environment, which is far from real-world scenarios. 
In reality, human activity is a continuous activity and there is no pre-set start time and end time. 
Thus, it is still an open question about how to automatically segment different activities and ensure the accuracy of continuous activity recognition. 

To address these challenges, we first extract sparse point cloud-based global embedding to represent the posture status of human activities at each frame, using the lightweight designed light-PointNet backbone~\cite{han2024mamba3d}. 
We then propose a bidirectional LiteLSTM (BiLiLSTM) model~\cite{elsayed2022litelstm} to the time-distributed frame embeddings to classify activities within a fixed time window. 
This concatenated architecture can explicitly extract the spatial and temporal features of human activities and conduct accurate human activity recognition. 
Furthermore, we combine our HAR model with transition optimization algorithms, including the Hidden Markov Model (HMM)~\cite{eddy1996hidden}
and Connectionist Temporal Classification (CTC)~\cite{graves2006connectionist}, to enhance the accuracy and robustness of real-time continuous HAR in real-world application scenarios. 

Specifically, we adopt the following steps in this study. 
First, we collect 3D point cloud data from a public dataset called MMActivity~\cite{singh2019radhar}, as well as from two self-collected datasets: discHAR (discrete data) and contHAR (continuous data).
Second, we develop a segment-wise point cloud augmentation (SPCA) algorithm, which includes data preprocessing, hybrid alignment of point cloud data, and segment-wise augmentation. 
Thirdly, we extract the point cloud global embedding using Lite-PointNet backbone on the sparse and imbalanced data. 
Cumulative features over time steps are then fed into the BiLiLSTM network for feature learning and activity classification. Finally, we apply an integrated transition optimization algorithm, combining HMM and CTC, to enhance the robustness and accuracy of continuous HAR. 
In particular, we have creatively ported our model to the Raspberry Pi which is connected with the mmWave radar and mounted on a movable healthcare monitoring platform. This forms a low-cost and deployment-friendly healthcare monitoring robot that can be used in real-world scenarios. 
Based on extensive experiments, our results show that our method performs accurately, efficiently, and robustly in both discrete and continuous HAR scenarios.

This paper contributes to the literature as follows.

\par
1) To the best of our knowledge, this study marks the first to develop a health monitoring robot using mmWave radar-based HAR systems, named RobHAR. 
It achieves mobile, lightweight, continuous HAR in real-world scenarios, addressing challenges such as handling sparse point cloud-based HAR, enabling real-time prediction on edge computing platforms, and overcoming fixed monitoring ranges.

2) A sparse point cloud-based global embedding using Light-PointNet (LPN) backbone is proposed to learn the point cloud feature. 
Experiments show that the global embedding contains sufficient spatial information to represent human activities in each frame, even in the case of very sparse and unbalanced point clouds.
SPCA, a segment-wise point cloud augmentation technique, is also proposed to improve the quantity and quality of the sparse point cloud data, reducing the workload of labeling and ultimately significantly improving the robustness and accuracy of the model. 

\par
3) A bidirectional lightweight LSTM model, BiLiLSTM, is proposed to learn the temporal pattern of human activities, 
The time-distributed global features are accumulated by consecutive time steps to represent the spatio-temporal pattern of different human activities. 
Furthermore, a transition optimization strategy is proposed on the HAR model to enhance the robustness of continuous HAR. By integrating HMM with CTC, this transition optimization method efficiently reduces false alarms in continuous HAR and obtains satisfactory accuracy and stability. 
\par
4) Extensive experiments have been conducted on two discrete and one continuous HAR datasets. The results demonstrate that our proposed LPN-BiLiLSTM model significantly outperforms the benchmarks, achieving higher accuracy with lower computing costs on all datasets. Additionally, our proposed transition optimization strategy, HMM-CTC, further enhances the robustness and accuracy of continuous HAR. Particularly, our system is deployed on a robot-mounted mmWave radar-based edge computing platform, achieving accurate and continuing healthcare monitoring in real-world scenarios. 

The remainder of this paper is organized as follows. 
Section~\ref{sect:relatedwork} briefly reviews the related work of point cloud feature representation, HAR models and transition optimization algorithms. 
Section~\ref{sec: RobHAR} introduces our RobHAR system, which consists of segment-wise point cloud augmentation, sparse point cloud global embedding, HAR model, and transition optimization. 
Section~\ref{sec: results} presents the experiment settings and the evaluation of our proposed approach, with extensive experiments on three datasets. 
Section~\ref{sec: conclusion} concludes the paper.

\section{Related Work}
\label{sect:relatedwork}

Intensive research works have been conducted on radar-based HAR in recent years. 
This paper focuses on sparse point cloud-based HAR on mmWave radar. 
In this section, we focus on sparse point cloud feature representation, HAR models, and transition optimization strategies for HAR. 

\subsection{Point Cloud Feature Representation}

Generally, mmWave radar generates time series signals, and after signal transformation, it can output a series of frames. 
Each frame comprises a random number of points. In the HAR area, feature extraction is based on windowed data (\textit{i.e.}, data within each time window segment) to produce feature vectors for recognition. 

The earlier studies normally extract point cloud features with generic volumetric feature extraction methods by voxelating sparse point cloud into a 3D space~\cite{singh2019radhar,zhao2019mid}. 
Sing \textit{et al.}~\cite{singh2019radhar} proposed RadHAR framework to classify five human activities through a mmWave radar. 
By voxelating, they generated a $60 \times 10 \times 32 \times 32$ (614,400) feature vector as an input to classifiers. Zhao \textit{et al.}~\cite{zhao2019mid} proposed a human tracking and identification system (mID) to identify people’s unique characteristics based on mmWave radar. They created an occupancy grid cube and then flatted it into a feature vector of a dimension of 16,000 for each frame.
The feature volume of the mID is much larger ($16,000 \times 33 \times 2=1,056,000$) than RadHAR. 
The main limitation of the above-mentioned volumetric methods is enormous memory occupation (\textit{e.g.}, RadHAR requires more than 45GB for training and 16GB for testing while mID requires an even larger occupation) and computational intensity in training and testing~\cite{zhou2018voxelnet}.

A breakthrough made by Qi \textit{et al.}~\cite{qi2017pointnet} is that PointNet was introduced to overcome the drawbacks of volumetric methods. 
PointNet is an end-to-end deep neural network that directly consumes point clouds to learn point-wise and global features. Extensive empirical evidence shows that this approach enables the network to sufficiently represent the global features of point clouds in a broad range of application scenes, such as object classification and part segmentation. 
A series of following research further extends this method~\cite{qi2017pointnet++,li2018pointcnn,wang2019dynamic,liu2024point, han2024mamba3d,saydam2023feature}. 
They mainly focus on improving local structure representation in object classification and part segmentation.  

Later, inspired by the Transformer~\cite{vaswani2017attention} models, various iterations such as Point Transformer v1-v3~\cite{zhao2021point, wu2022point, wu2023point} have gained attention and demonstrated cutting-edge performance. 
However, despite their success, Transformer-based backbones require substantial computational resources and encounter challenges in quadratic complexity and handling long sequences~\cite{han2024mamba3d}.

Recently, Mamba~\cite{gu2023mamba} empowered point cloud representations such as PointMamba~\cite{liu2024point} and Mamba3D~\cite{han2024mamba3d}, have shown superior performance and higher efficiency than the Transformer-based models, using a light-PointNet model to extract patch embedding for small amount of points, integrated with a state space model to represent the overall large amount of points. Inspired by their success and considering the limited computational power of our robot platform, as well as the sparsity of our point clouds generated by mmWave radar, we select only the patch embedding model light-PointNet (LPN) as backbone to extract the point cloud feature.

\subsection{HAR Models}

In early studies, machine learning techniques such as Support Vector Machine (SVM), K-Nearest Neighbor (KNN), Decision Tree (DT), Quadratic Discriminant Analysis (QDA), Random Forest (RF) have been widely used in small-scale static HAR scenarios~\cite{xu2019lecture2note, li2019survey}. 
Later, an increasing number of deep learning methods including convolutional neural networks (CNNs), recurrent neural networks (RNNs), Gated Recurrent Units (GRU), and Auto-Encoder (AE), and have demonstrated superior performance with the conventional machine learning methods~\cite{xu2022morphtext,xu2024seeing,singh2019radhar}.  

In recent years, with the success of Transformer~\cite{vaswani2017attention}, there has been a surge in time series modeling methods based on them, including Informer~\cite{haoyietal-informer-2021}, Dliner~\cite{zeng2023transformers}, and SCINet~\cite{liu2022SCINet}. 
These models have demonstrated state-of-the-art performance in time series data prediction tasks, such as HAR. 
However, Transformer-based models incur high computational costs and are not suitable for direct deployment on robot-mounted edge devices for real-time HAR~\cite{han2024mamba3d}. 
Recent research by Xie \textit{et al.}~\cite{xie2024tsc} has shown that a concatenate model, which combines feature embedding with the classic MLSTM-FCN model, can achieve comparable performance compared to the Transformer-based models. 
Additionally, a light-weighted LSTM model LiteLSTM~\cite{elsayed2022litelstm}, is proven to be more accurate and efficient than the traditional LSTM model in HAR task.

Inspired by these previous works, in this study, we propose a concatenate model for HAR. It stacks the LPN-generated point cloud embeddings by consecutive time steps, then combines a bidirectional LiteBiLSTM (BiLiLSTM) model to classify human activities accurately. 

\subsection{Transition Optimization}
Although numerous successful HAR research works have appeared in recent years, most of them are based on segmented time sequences for learning and predicting human activities, without considering the fact that there is no pre-segmentation in real-world application scenarios. 

Researchers have conducted exploratory studies on transition optimization. Ding et al.~\cite{ding2019continuous} proposed a concept of transitions that refers to a timeslot between two meaningful activities. They introduced a peak search method based on dynamic range-Doppler trajectory to extract the local maximum of Doppler components and select the most appropriate window centered on the peak points. 
Coppola et al.~\cite{coppola2019social} applied an HMM to learn a transition probability distribution between two activity states from the number of state changes in the training set. The HMM is capable of enhancing the robustness and accuracy of segmentation estimation by eliminating potential errors. 
Another method dealing with transitions between valid motions is Connectionist Temporal Classification (CTC)~\cite{graves2006connectionist}. 
CTC is widely used in the field of speech and text recognition, and it has been shown to be effective in filtering the transition gap between adjacent words~\cite{xu2022arbitrary, xu2022semantic}.
Zhang et al.~\cite{zhang2018latern} implemented the CTC algorithm in the field of dynamic continuous hand gesture recognition using radar sensors. They demonstrated that CTC performs effectively in processing unsegmented human activity sequences. 

In this study, we design a joint architecture that synthetically integrates HMM and CTC with the proposed HAR model, which obtains continuous, accurate and real-time HAR in the real world.
\begin{figure*}
    \centering
    \includegraphics[scale=0.32]{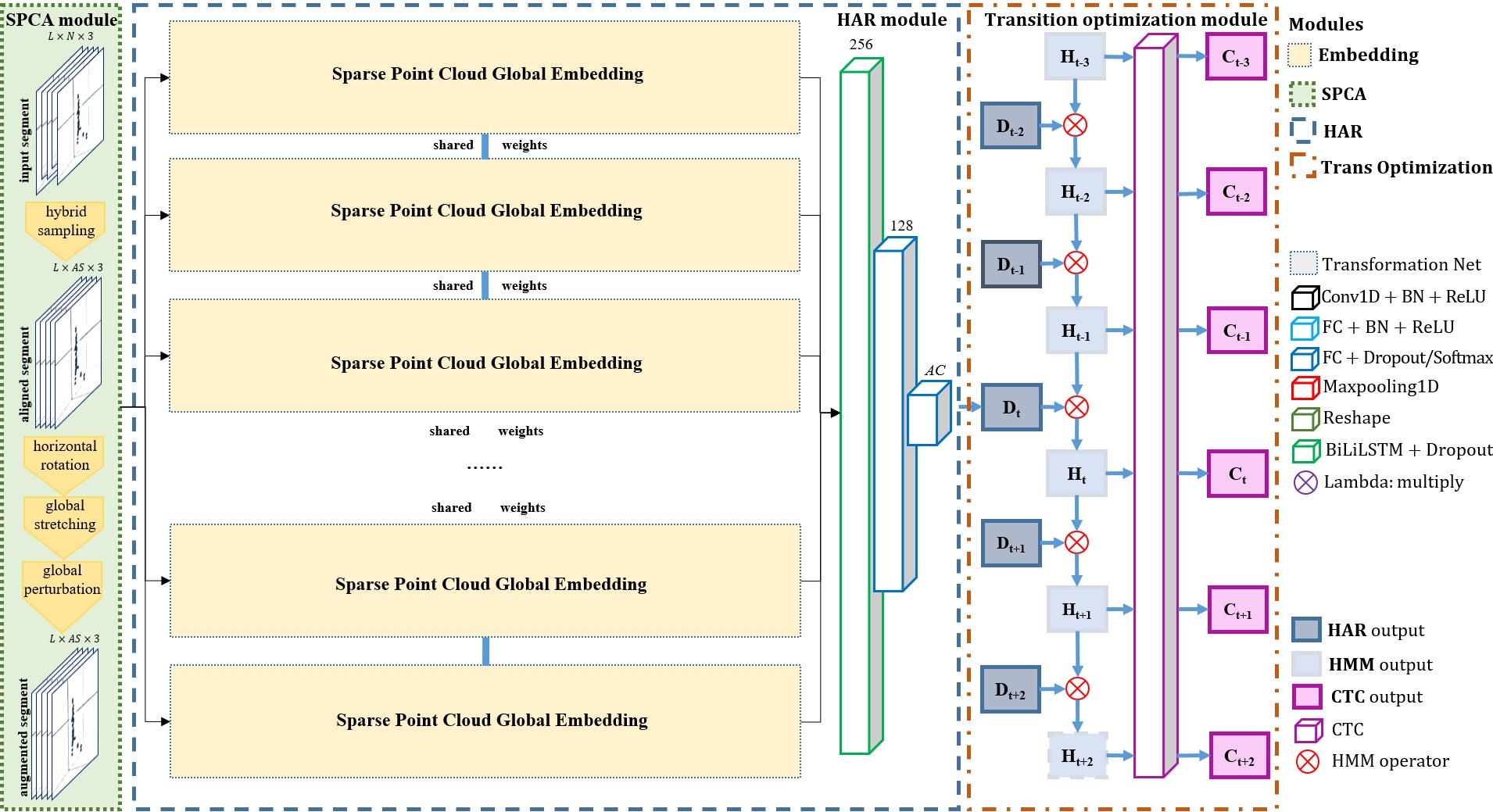}
    \caption{The architecture of our proposed RobHAR system.}
    \label{fig_RobHAR}
\end{figure*}

\section{RobHAR System}
\label{sec: RobHAR}
In this section, we provide a detailed explanation of our proposed RobHAR system, which is comprised of four modules, including SPCA, sparse point cloud global embedding, HAR, and transition optimization module. 
The architecture of RobHAR is shown in Figure~\ref{fig_RobHAR}.

\subsection{Segment-wise Point Cloud Augmentation}
\label{sec: SPCA}

The task in this study is classifying human activities based on a sparse point cloud generated by a mmWave radar. 
For this purpose, we extract global embedding using LPN model, which directly learns the features of sparse and non-uniformed point clouds. 
To ensure a balanced input and enhance the generalization of the point cloud global embedding, we develop a segment-wise point cloud augmentation (SPCA) algorithm, as detailed below.

\subsubsection{Hybrid Alignment}

Given non-uniformed raw point clouds from mmWave radar, we adopt a hybrid alignment method, \textit{i.e.}, hybrid sampling to ensure a balanced input fed to the LPN model. 
We denote a set of point clouds as $PC$, where each point 
has coordinates of ($x, y, z$) in the radar polar coordinates. 
The hybrid sampling method automatically implements upsampling or downsampling when the number of points in each frame is smaller or larger than the alignment size $AS$. Upsampling randomly replicates points from the frame until the total number of points reaches $AS$ while downsampling randomly selects the number of points equalling $AS$. 

To obtain the optimal $AS$, we compare and evaluate the impact of different $AS$ on the classification accuracy according to the distributions of point clouds on the experiment datasets, including MMActivity~\cite{singh2019radhar} and our constructed discHAR and contHAR dataset. For example, the MMActivity has a maximum of 25 points for each frame, while discHAR and contHAR have a maximum of 64 points per frame. Therefore, for MMActivity data alignment, we set $AS$ to 5, 10, 15, 20, and 25, respectively, to identify the optimal $AS$; 
for discHAR and contHAR, we set 10, 20, 30, 40, 50, and 64, respectively, to seek the optimal $AS$.

\subsubsection{Segment-wise Augmentation}

To enlarge the training sample size and enhance the robustness of our model, we develop the SPCA algorithm (see Figure~\ref{fig_SPCA} in Appendix~\ref{apsec: SPCA}), including segment-wise rotating, stretching and perturbating for the training set. 
Previous studies have successfully applied frame-wise augmentation methods in static object classification and part segmentation tasks~\cite{qi2017pointnet}. It is capable of augmenting the points randomly to any direction by any angle in 3D coordinates. 
To transfer it to our dynamic point clouds scenario, and based on the observation of human activities in the real world, we make the following two assumptions:

\textit{\textbf{Assumption 1.} Conventional frame-wise point cloud augmentation methods may easily fail in the HAR domain.}

\textit{\textbf{Assumption 2.}  Segment-wise augmentation will only take effect if it is consistent with the natural characteristics of human activities. 
Specifically, rotation can only be applied on the horizontal plane, stretching can be applied in the horizontal plane and vertical directions, and perturbation can be applied on all points.}

It is easy to prove Assumption 1 since numerous examples can be found in the real world. For example, when we rotate the points in a ``standing'' sample by 90 degrees, the points mistakenly represent ``lying'' after the rotation, thus misleading the learning model to classify the actual human activities.
For Assumption 2, it is also explainable. Rotating and stretching the segment (sequence of point cloud frames) on the horizontal plane is consistent with the fact that humans can move in various directions and at different speeds on the ground. Meanwhile, stretching in the vertical directions actually indicates higher or shorter objects. Perturbation indicates the slight movement of the human body, which is consistent with the phenomenon that our body keeps moving gently with the heartbeat and breath.

\textbf{SPCA formulation.} 
Figure~\ref{fig_SPCA} shows the detailed SPCA algorithm. 
We denote $P=\{P_{ij}|i=1,2,\cdots,L;j=1,2,\cdots,AS\}$ as a training segment with a shape of $(L, AS, 3)$,  
where $P_{ij}=(x_{ij},y_{ij},z_{ij})$ denotes the $j_{th}$ point in the $i_{th}$ frame in $P$; $L$ refers to time-window size, indicating that a training sample $P$ comprises $L$ consecutive frames; 
the alignment size $AS$ is related to the hybrid sampling methods discussed above, ensuring a balanced point cloud vector for the feature extraction model; and 3 indicates the 3D dimension of $xyz$ space. 
We further denote rotation angle $\theta$, stretching index $s$, and perturbation value $p$ for the SPCA. 

We first calculate the centroid $P_c=(x_c,y_c,z_c)$ of the point segment $P$. 
Then, we apply rotation on the horizontal plane by angle $\theta$ around the axis $u=(x_c,y_c,z_c)$. 
This is implemented by first mapping the points in $P$ to $P^M$ so that rotating around the axis $u'=(0,0,1)$ is implemented by straightforwardly multiplying a basic rotation matrix on the vertical axis. 
After that, a reverse operation is applied to update the $P^R$ back to the original coordinates. 
Then, the stretched points $P^S$ are obtained by multiplying the stretching index $s$ to all points and followed by adding the perturbation values $p=(p_x,p_y,p_z)$  to all points. 

Thus, the SPCA is successfully implemented on the training set to get augmented segment $P^A$. 
By repeating this operation by epochs, it can acquire tens of times larger number of training samples that represent a variety of human characteristics (tall or short, fat or slim), various moving directions (e.g. forward, backward, sideways), and different acting speeds (fast or slow).

\subsection{Sparse Point Cloud global embedding}
\label{sec:SPCGE}
This section introduces the details of sparse point cloud global embedding that extracts frame-wise features from sparse point cloud, using LPN model as the backbone. 

\begin{figure}
    \centering
    \includegraphics[scale=0.175]{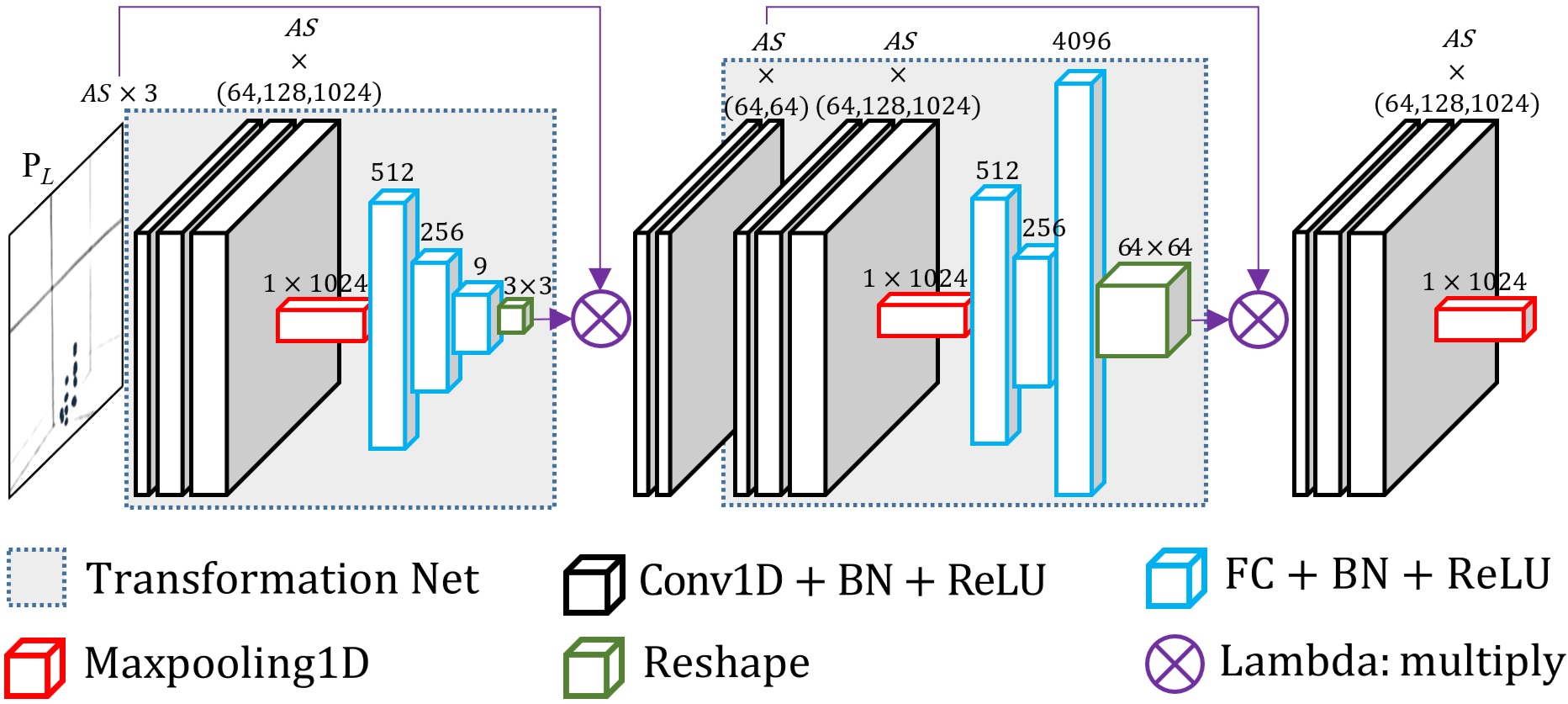}
    \caption{The sparse point cloud global embedding model to learn the frame-wise point cloud feature.}
    \label{fig_Embedding}
\end{figure}

As shown in Figure~\ref{fig_Embedding}, the feature extraction process begins with a Transformation Net, which standardizes the input point clouds, ensuring robustness against various transformations.

Subsequent layers of Conv1D + BN + ReLU are employed for feature learning. These layers, with batch normalization and ReLU activation, introduce non-linearity and efficiently capture essential features, progressively reducing dimensionality.
Maxpooling1D is then utilized to further abstract the features by taking the maximum value over intervals, significantly reducing spatial dimensions and emphasizing critical aspects. 
A Reshape layer follows, adjusting the dimensions to maintain compatibility throughout the architecture. This is crucial for the seamless flow of data between different stages.

Finally, the FC + BN + ReLU layer processes the reshaped features to learn complex patterns. The Lambda: multiply operation at this stage combines the features to generate a comprehensive global embedding that encapsulates the essential characteristics of the original sparse point clouds.

This structure exemplifies a systematic approach to handling specialized data, ensuring effective feature extraction through a series of transformational, convolutional, pooling, reshaping, and fully connected operations.

\subsection{HAR module}
In this section, we employ the bidirectional lightweight LSTM (BiLiLSTM) as the HAR module, as shown in Figure~\ref{fig_RobHAR}. 

The HAR module is designed for human activity pattern learning, utilizing a time series of point clouds as raw data. 
As we described in the previous section, each frame of point clouds is extracted through the sparse point cloud global embedding module first, and shared weights are applied to extract significant features from the point clouds. 
These features are then processed through a BiLiLSTM network with 256 units, which captures temporal patterns by analyzing data in both forward and backward directions. This comprehensive temporal analysis allows for a nuanced understanding of the sequence of activities. 
The output from the BiLiLSTM is condensed to 128 units before reaching the Activity Classification (AC) layer, which classifies the activities into distinct categories based on the learned temporal patterns. 

This architecture highlights the potential for real-time, efficient human activity recognition from point cloud data.

\subsection{Transition Optimization}
To achieve continuous real-time HAR, a popular approach is to split the time series data into sliding windows~\cite{morrison2020representing}. 
However, there is the potential for windows to overlap and not correctly represent realistic activity behaviors~\cite{coppola2019social}.

In our RobHAR system, the HAR module independently outputs a series of classification results without considering the relationship between adjacent predictions. This contradicts the continuous nature of human activities, which exhibit close relationships between adjacent time steps.

In this section, we introduce a transition optimization strategy that employs a combination of HMM and CTC methods to enhance performance in continuous HAR tasks. This dual approach is designed to improve accuracy and avoid abnormal predictions.
In the HMM stage, the hidden states $Y$ is the ground truth of human activities, the observed data $D$ denotes the outputs of the HAR module, and the observed states $X$ indicate the predicted activities. 
Based on the labeled training dataset, the HMM model learns parameters including 1) start probability $\pi$, represents the initial likelihood of each state $Y_i$; 2) emission probability $A$, denotes the probability of $Y_i$ generating $X$; 3) transition probability $B$, refers to the chance of transitioning from one state $Y_i$ to another $Y_j$. 
The output $H$ encapsulates these predictions, providing an optimized probability to better interpret the pattern of human activities.

The CTC stage introduces a blank placeholder $\epsilon$ to handle alignment issues between input sequences and output labels during training. This stage refines the activity recognition results by filtering out the blank transition frames between two adjacent activities. At last, it produces the final optimized output $C$.

By integrating HMM’s temporal dependency modeling with CTC’s sequence alignment robustness, our proposed transition optimization strategy ensures a more accurate representation of transitions between different human activities, which is crucial for the reliability of continuous HAR systems. This sophisticated approach highlights the potential for advanced applications in real-time healthcare monitoring and analysis.

\section{Experiments and Evaluation}
\label{sec: results}
In this section, we first introduce the experiment settings, including the details of datasets, the comparative algorithms, and the evaluation metrics. Then, we explain the details of the selection of time window size, the evaluation of SPCA, HAR, transition optimization and the computational costs.

\subsection{Experiment Settings}
\subsubsection{Datasets}
We evaluate the performance of our RobHAR on three datasets, MMActivity~\cite{singh2019radhar}, and two self-constructed datasets, including a discrete dataset discHAR and a continuous dataset contHAR. 

\textbf{MMActivity}. The MMActivity Dataset\cite{singh2019radhar} is an indoor human activity dataset consisting of 5 activities, including walking, jumping, jumping jacks, squats, and boxing. 
Each activity is performed by two subjects in front of the radar, which is placed on a tripod at a height of 1.3m in a laboratory environment. 
The radar chip is from TI’s low-cost commercial product IWR1443BOOST. It has three transmitters and four receiver antennas, making it capable of detecting moving objects in the 3D plane and producing point clouds with multi-dimensional information, including spatial coordinates, velocity, range, intensity and azimuth. 
The entire dataset is acquired at a 30Hz sampling frequency, with a duration of around 20 minutes for each activity and 93 minutes in total.

\textbf{DiscHAR and ContHAR}. 
The discHAR and contHAR datasets are acquired by a movable robot-mounted mmWave radar platform. It has 3 TX antennas and 4 RX antennas, where the radar frame period is 100 ms. 
The chirp start frequency is 77 GHz, and 48 chirp signals are issued during each frame period.
The sweep bandwidth is 3.2 GHz, and the chirp sweep rate is 100 MHz/us. 
The radar installation height is about 1.6m, the pitch angle is about 20 degrees, and the active test area of $3.5m \times 3.5m$ is marked 1.2m ahead of the radar. 
A total of 14 sets of data sets are collected in the experiment. 
The total duration of the data sets is about 100 minutes. Five human activities are labeled in the dataset, including walking, falling, standing, rising and lying.

The discHAR dataset collects each independent individual activity without considering the transition between two events. 
For example, the samples of ``walking" and ``falling" are collected separately. 
The contHAR dataset collects continuous events that are closer to real-world scenarios. 
For example, a subject sequentially performs walking followed by falling and then lying on the floor. 
Thus, this dataset collects both activity frames and the transition frames between two activities. 

\begin{figure*}[ht]
    \centering
    \includegraphics[scale=0.385]{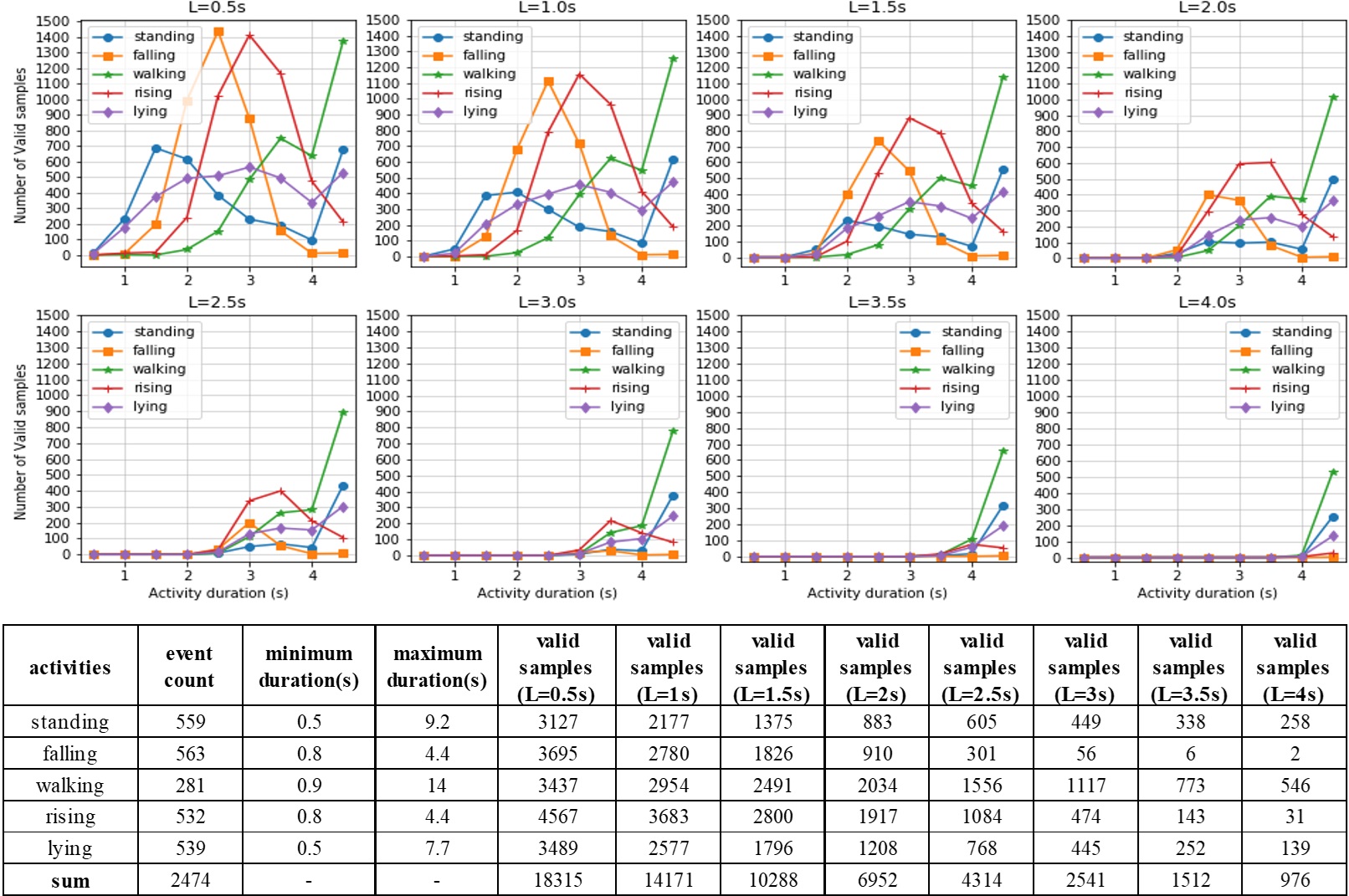}
    \caption{The statistics of valid samples of the discHAR and contHAR dataset. The $L$ denotes the size of the time window. With different $L$ values, the valid samples in both datasets change dramatically.}
    \label{fig_Samples}
\end{figure*}

The number of valid samples for training is mainly corresponding to the size of the time window $L$. 
As shown in Figure~\ref{fig_Samples}, the total activity events count for 2,474 while each event lasts for different durations, from 0.5 seconds to 14 seconds. Hence, the number of valid samples varies from 18,315 to 976. 
We have conducted comprehensive experiments to select the optimal time window size $L$ in below sections.

\subsubsection{Comparative Algorithms}
\label{sec: compAlgorithoms}

We compare the benchmark models and our proposed models in our experiments. 

\textbf{TD-CNN-BiLSTM}~\cite{singh2019radhar}. This is the baseline model of the MMActivity dataset, using the TD-CNN model to extract the point cloud feature and a BiLSTM model as the classifier. 

\textbf{TD-CNN-MLP}~\cite{singh2019radhar}. This is a comparative method proposed in the MMActivity dataset that applies the MLP network for the classification of human activities. 

\textbf{LPN-GRU}. This is a concatenate model developed by us using the LPN as a feature extraction model and a GRU as the human activity classifier. 

\textbf{LPN-BiLiLSTM}. This is a concatenate model proposed by us using the LPN model to extract the global embedding of sparse point clouds and using bidirectional BiLSTM as a lightweight but efficient classifier for HAR. 

\subsubsection{Evaluation Metrics}
Our evaluation matrices contain the micro accuracy ($micA$), macro precision ($P$), macro recall ($R$) and macro F1 score ($F1$).
The micro accuracy refers to the proportion of correctly classified samples. 
The macro precision and recall are computed by averaging precision and recall for all classes. 
The macro F1 score is calculated by the macro precision and recall by $F1=2*P*R/(P+R)$.

\subsection{Selection of Time Window Size}

Human activity is composed of a series of continuous body motions over a period of time. 
It is a key issue to select an appropriate size of the time window for the HAR model. 

Intuitively, if a time window is too short, it is not sufficient to represent the entire activity, while if a time window is too long, it is not sensitive to speedily occurring activities. 
Based on the experience of previous studies, the baseline model TD-CNN-BiLSTM on MMActivity simplistically set a time window of 2 seconds with a sliding window of 0.33 seconds, achieving an accuracy level of 90.47\%. 
In our study, differently, to evaluate the impact of time-window size $L$ on the classification accuracy, we set $L$ to 0.5, 1.0, 1.5, 2.0, 2.5 and 3.0 seconds, respectively.

\begin{figure*}[h]
    \centering
    \includegraphics[width=6in,height=2.5in]{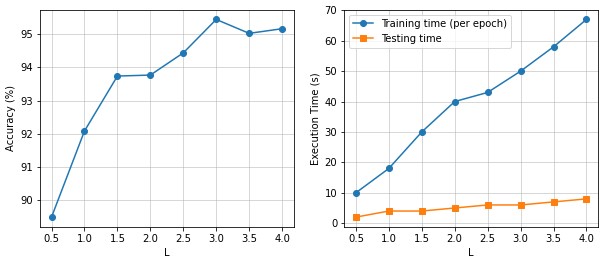}
    \caption{The impact of the time window $L$ on the accuracy and execution time on the MMActivity dataset}
    \label{fig_LA_MMActivity}
\end{figure*}

We use our proposed LPN-BiLiLSTM model on the MMActity dataset to evaluate and select the optimal time window size.
The left part in Figure \ref{fig_LA_MMActivity} shows that the classification accuracy basically increases as $L$ increases and remains stable after the highest point. Specifically, the classification accuracy improves from 89.48\% when $L$ is set to 0.5, reaches the highest point 95.45\% when $L$ is set to 3, and keeps accuracy over 95\% when $L$ is set to longer than 3, which is significantly superior than the MMActivity benchmark. 
This indicates that too-short time windows fail to cover the entire event, while after the accuracy reaches the highest point, the additionally added time window cannot make incremental contributions to accuracy enhancement. 
It is worth mentioning that the accuracy of our model achieves over 93\% when $L$ is set to 2, suggesting that, compared to the RadHAR benchmark, our LPN-BiLiLSTM model achieves a higher accuracy level at the same settings. 

The right part in Figure~\ref{fig_LA_MMActivity} shows the testing and training time linearly increase as the length of the time window increases. 
Specifically, training time increases from 10 seconds to 67 seconds per epoch when $L$ is set from 0.5 to 4. Testing time increases from 2 seconds to 8 seconds when $L$ is set from 0.5 to 4. 
Thus, Figure~\ref{fig_LA_MMActivity} and~\ref{fig_LA_MMActivity} collectively indicate that if our goal is to achieve the optimal accuracy level, we should set $L$ to 3, while if our goal is to obtain optimal sensitiveness and ensure a satisfactory accuracy (around 94\%), we should set $L$ to 2.  

\subsection{Evaluation of SPCA}

To ensure a uniform input to the deep neural networks, we propose the segment-wise point cloud augmentation algorithm SPCA, including the hybrid alignment and augmentation of the sparse point clouds. The augmentation is automatically applied during the training process, while the hybrid alignment requires a prior definition of the number of points per frame for sampling.

The mainstream of point cloud datasets is generated by Lidar or depth camera, which comprises an extensive number of points in each frame (approximately from ~1k to 100k)~\cite{zhou2018voxelnet}. 
Thus, downsampling is usually applied on the dense point clouds. 
For example, PointNet downsampling the non-uniformed points by randomly selecting 2,048 points for each target~\cite{qi2017pointnet}. 

However, the point clouds generated by a mmWave radar in this study are highly sparse and imbalanced. 
Simplistically utilizing downsampling may not be the best choice here. 
We thus evaluate the performance of different point cloud sampling methods to show the impact on classification accuracy. 

\begin{figure*}[h]
    \centering
    \includegraphics[width=6.5in,height=2in]{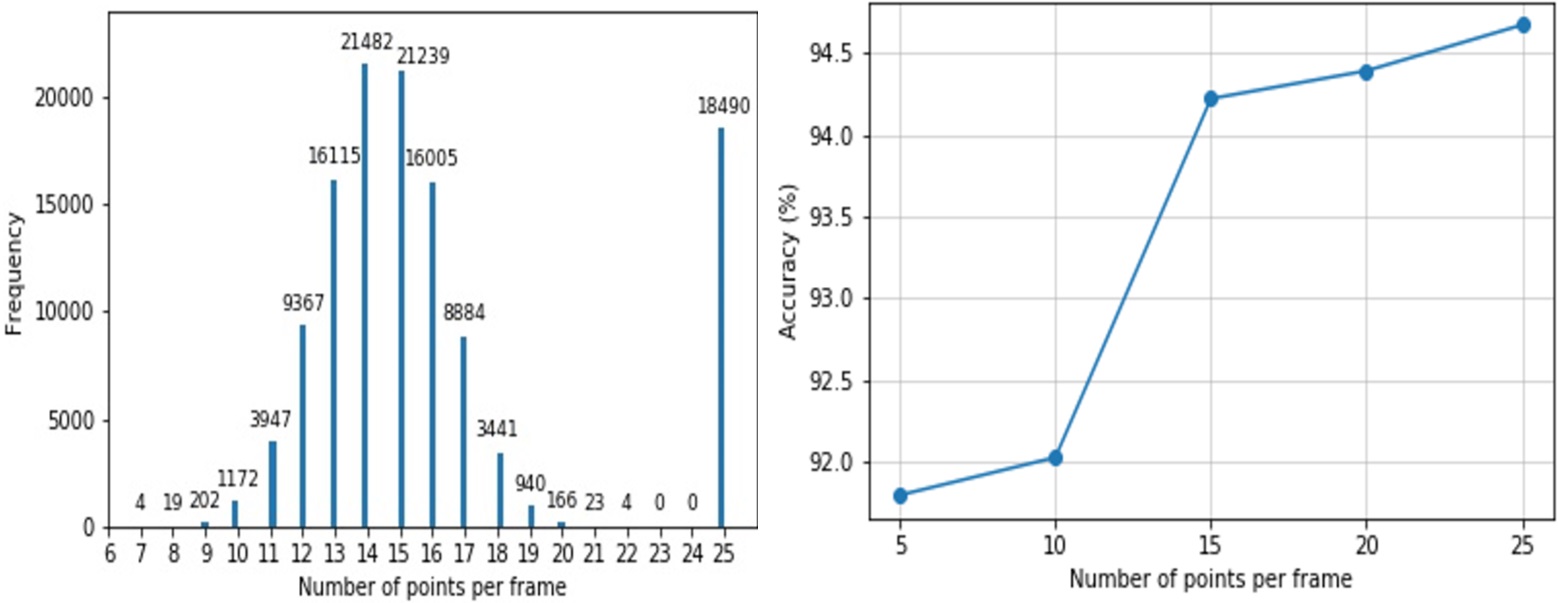}
    \caption{The statistics and impact of number of points per frame on the MMActivity dataset}
    \label{fig_SPCA evaluation}
\end{figure*}

For example, the mmWave radar in the MMActivity dataset works at a frequency of 30 frames per second. 
Figure~\ref{fig_SPCA evaluation} shows the statistical distribution of the number of points in each frame. 
Each frame has a minimum of 7 (when the object is moving slightly or even keeps static) and a maximum of 25 points (when the object is moving strongly). 
The frames containing 25 points account for around 15\% of total training samples. 
Other samples roughly follow a normal distribution with the mean between 14 and 15, indicating that 14 or 15 points in each frame can mostly represent human activities.

We further investigate the impact of sampling methods on classification accuracy by setting the alignment size $AS$ to 5, 10, 15, 20, and 25, respectively. 
We adaptively use downsampling, upsampling and hybrid sampling methods according to the setting of $AS$. 
When $AS$ is set to 5, downsampling is applied to all frames, while when $AS$ is set to 25, upsampling is applied to all frames; when $AS$ is set to 10, 15, or 20, downsampling is applied to those frames with a larger number of points than $AS$, otherwise, upsampling is applied. 
The results in Figure~\ref{fig_SPCA evaluation} are generated using the LPN-BiLiLSTM model. 
Basically, classification accuracy increases when $AS$ increases. When $AS=25$, the accuracy reaches a peak of over 94.5\%. 
This indicates that our upsampling method effectively ensures the completeness of information, at the same time, achieves a high-level accuracy. 
When $AS$ is equal to 15 or 20, the accuracy is slightly lower than that at 25 while when $AS$ is equal to 5 or 10, the accuracy drops dramatically to lower than 92\%. 
This indicates that as $AS$ keeps reducing, we lose more and more information contained in original point clouds. However, it is worth noting that when $AS$ is set to 15, the accuracy still remains at a high level. 
Consistent with the above discussion, 15 points per frame is capable of representing the most features of activities. 

\subsection{Evaluation of HAR}

To perform a comprehensive and fair comparison of the HAR models on the three datasets, we apply two feature extraction algorithms TD-CNN and LPN, integrate with HAR models. Eventually, four concatenate models, including TD-CNN-MLP, TD-CNN-BiLSTM, LPN-GRU and LPN-BiLiLSTM are evaluated (see more details in Section~\ref{sec: compAlgorithoms}). 
In addition, we test the computational costs of all models. The results are shown in the tables below. For all four models on each dataset, the time window size $L$ is set to 2s, and the alignment size $AS$ is set to 25 on MMActivity and 64 for both discHAR and contHAR datasets.
In the tables, the ``Micro Accuracy" refers to micro accuracy. The ``Mac-P" refers to macro Precision, ``Mac-R" refers to macro recall and ``Mac-F1" refers to macro F1 score.

\subsubsection{Results on MMActivity}
The experiment results on MMActivity are displayed in Table~\ref{table:Classi_result_MMAct}. 
Our proposed methods all significantly outperform the comparative methods. 
Specifically, the LPN-BiLiLSTM model achieves the highest F1 score at 95.29\%. 
The LPN-GRU model obtains an F1 score of 94.21\%. Collectively, our proposed models all perform much better than the benchmark methods. 

\begin{table}[h]
\centering
\caption{HAR results on MMActivity dataset}
\label{table:Classi_result_MMAct}
{\begin{tabular}[c]{@{}lcccc}
\toprule
  Method &Micro Accuracy&Mac-P&Mac-R&Mac-F1\\
\midrule
  TD-CNN-MLP\cite{singh2019radhar} & 83.86 & 84.69 & 84.01 & 83.81\\
  TD-CNN-BiLSTM\cite{singh2019radhar} & 90.42 & 91.37 & 90.55 & 90.66 \\
  \midrule
  LPN-GRU & 94.05 & 94.60 & 94.10 & 94.21 \\
  LPN-BiLiLSTM & $\textbf{95.12}$ & $\textbf{95.85}$ & $\textbf{95.18}$ & $\textbf{95.29}$ \\
\bottomrule
\end{tabular}}
\end{table}

\subsubsection{Results on discHAR}

Table~\ref{table:Classi_result_discHAR} presents the experimental findings on the discHAR dataset, where our developed models demonstrate a clear superiority over the existing methods. The LPN-BiLiLSTM approach, in particular, stands out with an F1 score of 95.45\%, the highest among all models. Overall, our methods consistently surpass the established benchmarks in performance.
\begin{table}[h]
\centering
\caption{HAR results on discHAR dataset}
\label{table:Classi_result_discHAR}
{\begin{tabular}[l]{@{}lcccc}
\toprule
  Method &Micro Accuracy&Mac-P&Mac-R&Mac-F1\\
\midrule
  TD-CNN-MLP & 94.03 & 94.55 & 94.10 & 94.21 \\
  TD-CNN-BiLSTM & 93.15 & 94.01 & 93.37 & 93.41 \\
  \midrule
  LPN-GRU & 95.14 & 95.39 & 94.14 & 94.67 \\
  LPN-BiLiLSTM & $\textbf{96.14}$ & $\textbf{96.51}$ & $\textbf{94.73}$ & $\textbf{95.45}$ \\
\bottomrule
\end{tabular}}
\end{table}

\subsubsection{Results on contHAR}
The experiment results on contHAR dataset are displayed in Table~\ref{table:Classi_result_contHAR}. 
Consistent with the results on the discHAR dataset, our proposed LPN-BiLiLSTM model demonstrates the best F1 score of 81.02. 
However, the values dropped largely from the results on discHAR as shown in Table~\ref{table:Classi_result_discHAR}. 
It is because the contHAR dataset includes lots of transition frames between two different activities. These segment of frames has blank or unclear features that can't represent any of the labeled activities. This illustrates the challenges in real-time continuous HAR and motivates us for the transition optimization, as discussed below.

\begin{table}[h]
\caption{HAR results on contHAR dataset}
\label{table:Classi_result_contHAR}
\centering
{\begin{tabular}[l]{@{}lcccc}
\toprule
  Method &Micro Accuracy&Mac-P&Mac-R&Mac-F1\\
\midrule
  TD-CNN-MLP & 70.11 & 71.56 & 70.88 & 71.21 \\
  TD-CNN-BiLSTM & 73.48 & 74.28 & 72.61 & 73.44 \\
  \midrule
  LPN-GRU & 80.25 & 80.51 & 79.84 & 79.94 \\
  LPN-BiLiLSTM & \textbf{81.62} & \textbf{82.10} & \textbf{81.13} & \textbf{81.02} \\
\bottomrule
\end{tabular}}
\end{table}

\subsection{Evaluation of Transition Optimization}

Table~\ref{table:eval_trans_opt} shows the evaluation results of using HMM and CTC algorithms as transition optimization strategies to enhance the accuracy and robustness of our model in a real-time continuous HAR setting.
The compared methods include LPN-BiLiLSTM, LPN-BiLiLSTM-HMM, and LPN-BiLiLSTM-HMM-CTC, with the latter integrating both HMM and CTC algorithms. 
The results indicate that both HMM and CTC show significant improvement in performance (above 85 F1 score) compared with the baseline LPN-BiLiLSTM model (81.02 F1 score). 
The LPN-BiLSTM-HMM-CTC method outperforms the others, achieving the highest scores across all metrics, which suggests the superior effectiveness of this integrated transition optimization strategy HMM-CTC in optimizing transitions for accurate and robust activity recognition. 
\begin{table}[h]
\centering
\caption{Evaluation of transition optimization with HMM and CTC}
\label{table:eval_trans_opt}
{\begin{tabular}[l]{@{}lcccc}
\toprule
  Method &Micro P/R&Mac-P&Mac-R&Mac-F1\\
\midrule
  LPN-BiLiLSTM & 81.62 & 82.10 & 81.13 & 81.02 \\
  LPN-BiLiLSTM-HMM & 85.13 & 85.65 & 84.96 & 85.01 \\
  LPN-BiLiLSTM-HMM-CTC & $\textbf{86.03}$ & $\textbf{86.32}$ & $\textbf{85.79}$ & $\textbf{85.91}$ \\

\bottomrule
\end{tabular}}
\end{table}

\subsection{Comparison of Computational Costs}
We conduct a comprehensive comparison of the computational costs of our methods with comparative methods with the same time window and generating almost the same number of samples, to evaluate the applicability of our lightweight model on a robot-mounted edge device. 
Computational costs include memory occupancy for training and testing, the time consumption for feature extraction, the number of parameters, and the execution time. 
The results are summarized in Table~\ref{table:Compu_cost_MMAct}. 

Specifically, the memory occupancy of our models is less than 1\% of models used in RadHAR (i.e., TD-CNN-MLP and TD-CNN-BiLSTM). 
The time consumption for feature extraction of our models is around 100 seconds, but the time in RadHAR models is over 2 hours. 
Our lightweight model's parameters (79k) are much fewer than their counterparts. 
Finally, the execution time of our optimized model is 17 seconds, while the time of the optimal model of comparative models is more than 46 times that of ours.

\begin{table}[h]
\centering
\caption{Computational cost on MMActivity}
\begin{threeparttable}
\label{table:Compu_cost_MMAct}
{\begin{tabular}[l]{@{}lcccccc}
\toprule
  Method &MO-tr&MO-ts&FET&Para&ExT\\
\midrule
  TD-CNN-MLP\cite{singh2019radhar}&45GB&16GB&2.3h&39.3m&130/31s\\
  TD-CNN-BiLSTM\cite{singh2019radhar}&45GB&16GB&2.3h&0.3m&796/79s\\
\midrule
  LPN-GRU&0.4GB&0.1GB&1.7m&2.8m&40/7s\\
  LPN-BiLiLSTM&\textbf{0.4GB}&\textbf{0.1GB}&\textbf{1.7m}&\textbf{79.7k}&17/2s\\
\bottomrule
\end{tabular}}
\begin{tablenotes}
\small
\item The ``MO-tr'' refers to memory occupancy of training set; ``MO-ts'' refers to memory occupancy of the testing set; ``FET'' refers to feature extraction time; ``Para'' refers to the number of total parameters and ``ExT'' refers to the execution time. The values under ``ExT'' follow the format of ``training time per epoch''/``testing time''.
\end{tablenotes}
\end{threeparttable}
\end{table}

\section{Conclusion}
\label{sec: conclusion}
This study has creatively proposed RobHAR, a mmWave radar-based HAR system mounted on a movable edge device as a healthcare monitoring robot. This system first extracts the global embedding of point clouds using a light-PointNet as the backbone. Then, the time series of features are concatenated with a bidirectional lightweight LSTM (BiLiLSTM) model to learn the pattern of human activities and make the prediction. 
In order to enhance the stability and accuracy of this model in continuous HAR, the HAR model has been integrated with transition optimization strategies, including HMM and CTC algorithms. Extensive experiments have been conducted on three datasets to evaluate the performance of our model in both discrete and continuing scenarios. Collectively, our model outperforms the benchmark point cloud representations and HAR methods in sparse point cloud-based continuous HAR in terms of classification accuracy and efficiency. Finally, to test the applicability of our approach in the real world, our RobHAR system has been deployed to a movable edge computing platform, forming a flexible healthcare monitoring robot and obtaining satisfactory performance on efficient, stable and continuing HAR in real-wrold scenarios.


\bibliographystyle{IEEEtran}  
\bibliography{references}

\newpage
\appendix
\section*{Appendix}

The pseudo-code for the segment-wise point cloud augmentation (SPCA) algorithm is shown in Figure~\ref{fig_SPCA} below.
\label{apsec: SPCA}

\begin{figure}[h]
    \centering
    \includegraphics[scale=0.83]{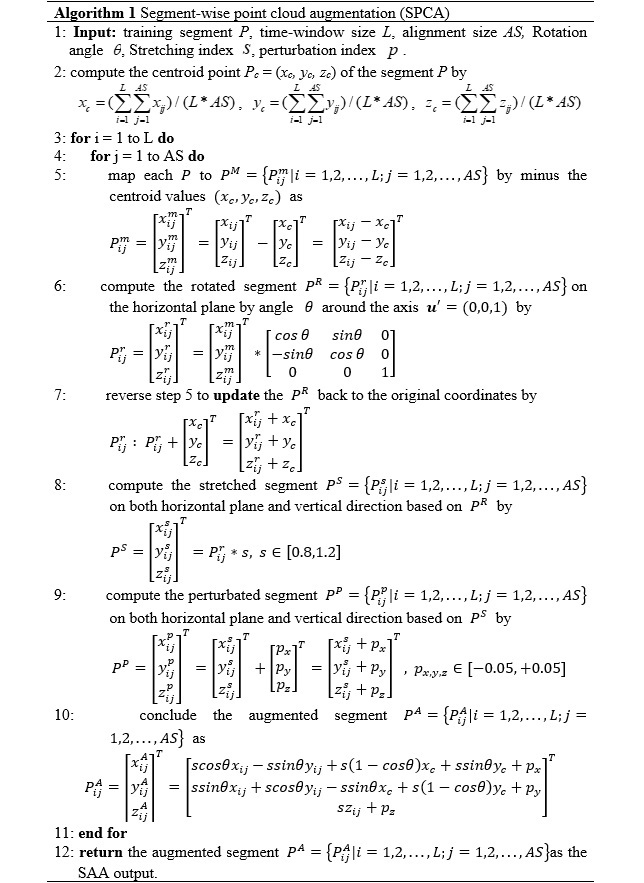}
    \caption{Segment-wise point cloud augmentation (SPCA) algorithm}
    \label{fig_SPCA}
\end{figure}

\end{document}